\documentclass{article} 
\usepackage{iclr2027_conference,times}

\usepackage{amsmath,amsfonts,bm}

\def\eqref#1{equation~\ref{#1}}

\def\1{\bm{1}}

\DeclareMathAlphabet{\mathsfit}{\encodingdefault}{\sfdefault}{m}{sl}
\SetMathAlphabet{\mathsfit}{bold}{\encodingdefault}{\sfdefault}{bx}{n}

\usepackage{hyperref}
\usepackage{url}
\usepackage{graphicx}
\usepackage{booktabs}
\usepackage{multirow}
\usepackage{wrapfig}

\title{Beyond the Encoder: Object-Centric Representations Emerging from Segmentation Foundation Models}

\author{Michalis Pistos \& Bart\l omiej W. Papie\.z \\
Big Data Institute, Li Ka Shing Centre for Health Information and Discovery, Nuffield Department of Population Health \\
University of Oxford \\
Oxford, OX3 7LF, United Kingdom \\
\texttt{michalis.pistos@ndph.ox.ac.uk, bartlomiej.papiez@bdi.ox.ac.uk}
}

\iclrfinalcopy 
\begin{document}

\maketitle

\begin{abstract}
Foundation models are increasingly used as feature extractors for downstream tasks with limited data, typically through representations obtained from their encoders. However, foundation models trained for structured tasks such as segmentation also form object-specific representations during task execution, whose transferability remains largely unexplored. We investigate Segment Anything Model 2 (SAM2) object pointers, compact object-level representations produced through the segmentation pathway, as transferable representations for downstream tumour prediction. Through experiments on two PET/CT tumour prediction tasks, we find that object pointers learned using segmentation supervision alone already encode tumour-discriminative information, despite never being optimised for the downstream labels. We further introduce task-guided object pointers, where an auxiliary objective jointly shapes the object pointer space for the downstream prediction together with segmentation. Task-guided pointers improve cancer-type classification, outperforming conventional encoder-based transfer and radiomics, while segmentation-only pointers perform best on the more challenging recurrence prediction task. Representation analyses reveal associations between object pointers and interpretable radiomics features, class-dependent organisation of the learned representation space, and explicit spatial grounding around the segmented tumour. Together, these findings suggest that useful transferable representations can emerge through the execution of a foundation model's primary task, providing an alternative to conventional encoder-based transfer.
\end{abstract}

\section{Introduction}
Medical images provide rich quantitative information~\citep{gillies2016radiomics} about tumour phenotype~\citep{fass2008imaging} that can support downstream tasks ranging from tumour characterisation to clinical outcomes prediction~\citep{erickson2017machine,bi2019artificial,koh2022artificial,jiang2023deep}. Learning representations that capture this information remains challenging, because task-specific imaging datasets are often small and clinical labels are limited. This motivates representations that capture informative tumour characteristics and can transfer across downstream objectives.

Several approaches have been explored to learn informative representations when medical imaging data are limited. Self-supervised learning methods leverage unlabelled data to learn transferable representations~\citep{rani2024self,shurrab2022self,wu2024voco}, while radiomics represent tumours through a compact set of predefined descriptors of shape, intensity, and texture~\citep{lambin2012radiomics}, and have been applied across a range of cancer applications~\citep{aerts2014decoding,yip2016applications,avanzo2020machine,liu2019applications}. More recently, foundation models offer representations learned from large and diverse datasets that can be transferred to data-scarce medical imaging tasks~\citep{moor2023foundation,ryu2025vision}. This has motivated both the adaptation of general-purpose foundation models and the development of domain-specific models for medical image segmentation, classification, and related tasks~\citep{ma2024segment,zhou2023foundation,xiang2025vision}.

However, downstream adaptation of foundation models commonly focuses on representations obtained from the model encoder, either by using it as a frozen feature extractor, or fine-tuning it for the target task~\citep{huang2025systematic,pai2024foundation,jacob2025towards,yang2025foundation}. 
Encoder features are often obtained before much of the computation associated with the model's primary task and therefore precede the task-specific processing that transforms image-level features into more specialised outputs. Representations emerging during this process may provide an alternative source of transferable information that is largely overlooked. This possibility is particularly interesting for segmentation foundation models~\citep{kirillov2023segment,zou2023segment,wang2023seggpt}. Segmentation requires the model to identify and delineate a particular object, introducing object-conditioned supervision that transforms generic image features into representation centred on the object of interest. Such representations may therefore encode object-centric characteristics beyond those strictly necessary for delineation. In cancer imaging, this raises a natural question: does the representation formed while segmenting a tumour also encode information useful for predicting its properties?

Segment Anything Model 2 (SAM2)~\citep{ravi2025sam}, a foundation model for image and video segmentation, provides a natural setting to investigate this hypothesis. In addition to spatial features, SAM2 produces compact \textit{object pointers} through its segmentation pathway to encode segmented objects across frames. Unlike image-encoder features, these representations emerge after prompt- and object-conditioned mask-decoder processing. We therefore investigate object pointers as transferable tumour representations and ask whether information learned through segmentation can support downstream prediction beyond the segmentation task.

We evaluate this hypothesis across two public FDG-PET/CT datasets and distinct downstream prediction tasks: cancer-type classification on FDG-PET-CT-Lesions~\citep{https://doi.org/10.7937/gkr0-xv29} and recurrence prediction on HECKTOR 2022~\citep{andrearczyk2023hecktor}. 
Our main contributions are:
\begin{itemize}
\item We identify SAM2 object pointers as an overlooked source of transferable object-level representations, showing that pointers learned through segmentation alone already encode information predictive of downstream targets.
\item We introduce an auxiliary objective for task-guided object pointer learning and show that downstream supervision can further improve these representations, while its benefit depends on the downstream task: task-guided pointers perform best for cancer-type classification, whereas segmentation-only pointers perform best for recurrence prediction.
\item  We characterise the information encoded by object pointers through radiomics correlation, representation-space organisation, and spatial analysis. We find associations with interpretable tumour characteristics and, for cancer-type classification, information complementary to handcrafted radiomics, alongside class-dependent structure and explicit spatial grounding around the tumour.
\item We show that preprocessing choices can substantially alter the downstream information retained by object pointers despite relatively similar segmentation performance, demonstrating that task performance alone does not fully characterise the quality of representations emerging through that task.
\end{itemize}

\section{Related Work}

\subsection{Foundation Model Representations for Medical Downstream Tasks} 

Foundation models are commonly adapted to downstream medical imaging tasks through representations obtained from their pretrained encoders, either by using them as frozen features or by fine-tuning the encoder for the target objective~\citep{paschali2025foundation}. This paradigm has been widely studied with general-purpose vision foundation models such as DINO and DINOv2~\citep{caron2021emerging,oquab2023dinov2}, whose encoder representations transfer effectively across medical imaging datasets and tasks~\citep{huang2024comparative,baharoon2023evaluating,prokop2023deep}. 

This encoder-centric approach has also been applied to segmentation foundation models. Comparative studies have benchmarked the SAM~\citep{kirillov2023segment} image encoder as a feature extractor alongside other foundation models for medical image classification~\citep{huix2024natural,wu2025rethinking}, while more targeted approaches have adapted SAM image-encoder representations for downstream tumour classification~\citep{gu2025adapting,sun2024medical}. These studies demonstrate that representations learned by foundation models can transfer effectively to medical downstream tasks, even when the model was originally developed for segmentation. 

However, these approaches largely consider representations obtained directly from the model encoder. For models trained on structured tasks such as segmentation, this motivates investigating representations that emerge beyond the encoder, after the model has performed task-specific processing, as a complementary source of transferable information.

\subsection{Segmentation-Guided Representations for Downstream Tasks}

Segmentation has also been used to introduce object-specific information into downstream representations. 
OCCAM~\citep{rubinstein2025we} uses segmentation to isolate objects in pixel space before independently encoding the resulting object-focused images for classification. In medical imaging, SAM-generated masks have been similarly used to construct lesion-focused representations for dermoscopic classification, guiding features from a separate ViT-based classifier~\citep{binzagr2026foundation}. SAM-IE~\citep{wang2024sam} further combines SAM-generated masks and contours with the original image to direct classification towards diagnostically relevant regions. These approaches use segmentation outputs to isolate or highlight the object of interest and subsequently inform downstream representation. In contrast, we investigate the object representation formed internally through the segmentation process itself, without constructing a separate representation from the predicted mask.

\section{Methodology}
\label{sec:method}

\subsection{SAM2 Object Pointers}

\textbf{Segment Anything Model 2.} SAM2 is a promptable segmentation foundation model for object segmentation in images and videos~\citep{ravi2025sam}. Given an input image, an image encoder first produces spatial features that are combined with information from previously processed frames through a memory attention. Together with user-provided prompts, these features are processed by the mask decoder to predict segmentation masks for the object of interest. SAM2 then stores information from each processed frame in a memory bank, allowing it to propagate object information across subsequent frames.

\textbf{Object Pointers.} In addition to spatial memory features, SAM2 stores a compact representation of each segmented object, termed as an \textit{object pointer}. Object pointers are derived from the output mask tokens of the mask decoder, which interact with image and prompt features during segmentation. The output token corresponding to the selected mask is projected to obtain the object pointer ($\mathbf{z}_{ptr}$),
\[
\mathbf{z}_{ptr} = f_{ptr}(\mathbf{o}_t) \in \mathbb{R}^{256},
\]
where $\mathbf{o}_t$ denotes the mask-decoder output token associated with the selected object mask, and $f_{ptr}$ is a learned projection. The resulting object pointer is stored in the memory bank alongside the spatial memory features and can subsequently be attended to when processing the object in later frames.

Unlike image-encoder features, the object pointers emerge after prompt- and object-conditioned segmentation processing and are associated with a specific segmented object. Although SAM2 uses these representations internally to support segmentation across frames, we investigate whether they also encode object characteristics that transfer to downstream tumour prediction.

\subsection{SAM2 Representations for Tumour Classification}
\label{subsec:tumour_cls}

\begin{figure}[t!]
\centering
\includegraphics[width=1.0\linewidth]{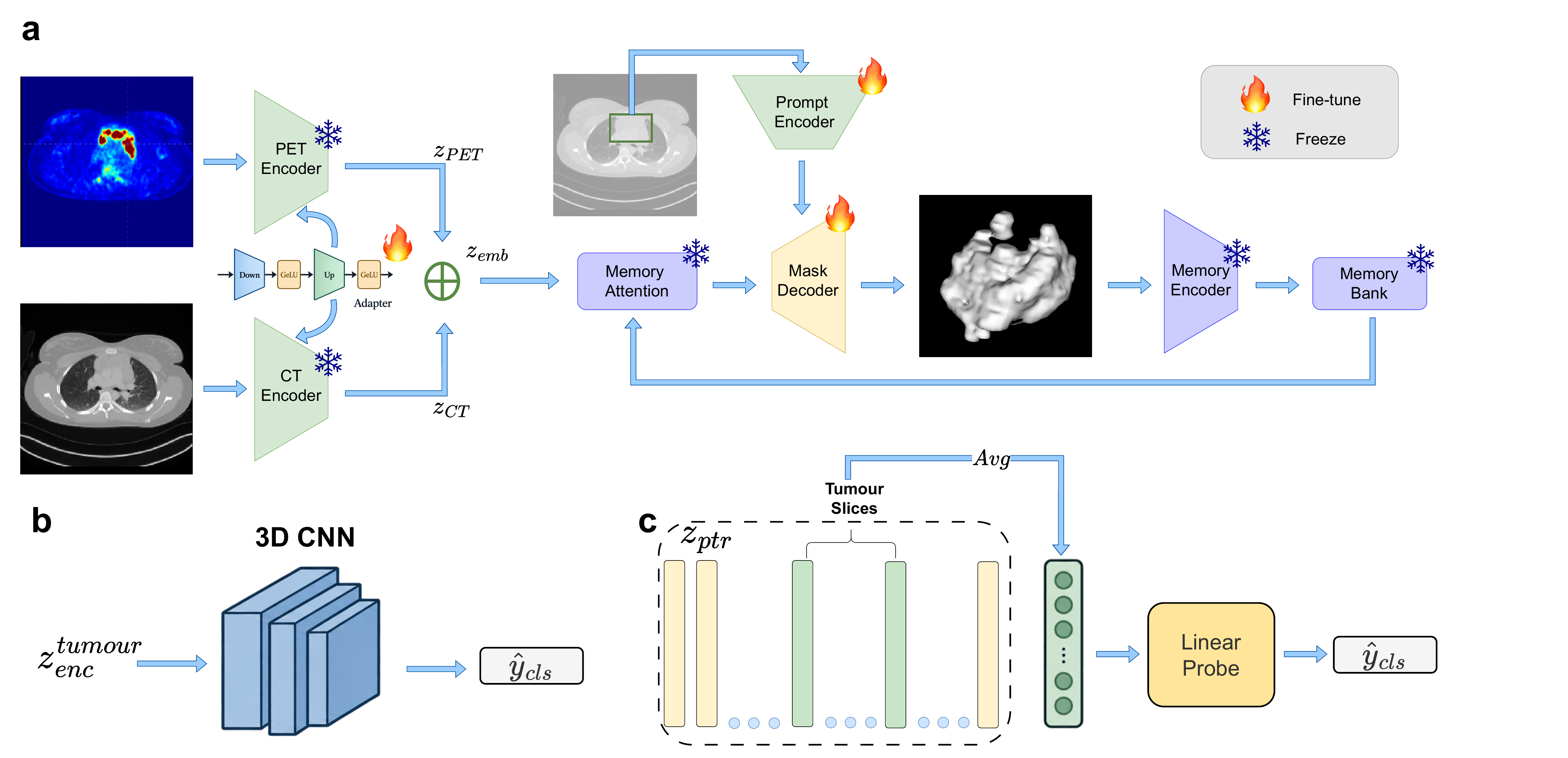}
\caption{Overview of SAM2 representations for downstream tumour prediction. 
(a) SAM2 adaptation to 3D PET/CT, using modality-specific encoders and a single bounding-box prompt for tumour segmentation. 
(b) Encoder-based transfer, where image-encoder features are used by a CNN for downstream classification. 
(c) Object pointer transfer, where slice-level object pointers produced through segmentation are aggregated into a tumour-level representation and classified using a lightweight linear head.
}
\label{fig:framework}
\end{figure}

\textbf{Problem definition.} Given a 3D PET/CT volume $\mathbf{x}$ containing a tumour, our objective is to obtain a tumour representation $\mathbf{z}$ that can be transferred to a downstream classification task. 
For each subject, the representation is used to predict a class label $y \in\{1,\ldots,C\}$, where $C$ denotes the number of classes for the corresponding downstream task. We formulate the downstream prediction as:
\[
\begin{gathered}
\hat{\mathbf{p}}_{cls}
= g_{\phi}\left(f_{\theta}(\mathbf{x})\right)
= g_{\phi}(\mathbf{z}),
\qquad
\hat{\mathbf{p}}_{cls} \in [0,1]^C, \\[2pt]
\hat{y}_{cls}
= \arg\max\left(\hat{\mathbf{p}}_{cls}\right),
\end{gathered}
\]
where $f_{\theta}$ denotes the representation function  that maps the input imaging volume $\mathbf{x}$ to a tumour representation $\mathbf{z}$, and $g_{\phi}$ is the downstream classification head producing the class probabilities $\hat{\mathbf{p}}_{cls}$. The final predicted class is denoted by $\hat{y}_{cls}$. We compare two sources of representations within SAM2: image-encoder features obtained before object-conditioned segmentation processing, and object pointers emerging through the segmentation pathway. This comparison allows us to assess whether the representations learned during SAM2's primary segmentation task encode transferable information beyond that captured by conventional encoder features.

\textbf{3D PET/CT segmentation.} We adapt SAM2 for 3D clinical PET/CT imaging by treating each volume as an ordered sequence of axial slices, analogous to video frames (see Fig.~\ref{fig:framework}a). Most pretrained SAM2 parameters are frozen, while the prompt encoder and mask decoder are fine-tuned, and lightweight trainable adapters are introduced within the image encoder for parameter-efficient adaptation~\citep{chen2024sam2}. To accommodate paired PET/CT input, we use separate modality-specific image encoders to obtain PET and CT representations, $\mathbf{z}_{PET}$ and $\mathbf{z}_{CT}$, which are fused using a learnable scalar $\alpha$,
\[
\mathbf{z}_{emb} = \alpha\mathbf{z}_{PET} + (1-\alpha)\mathbf{z}_{CT}.
\]
The fused representation is then passed to the SAM2 mask decoder for segmentation. A single perturbed bounding-box prompt derived from the ground-truth annotation is provided on the central axial tumour slice to specify the object of interest, with segmentation subsequently propagated through the volume without additional prompts. Further details on architecture, prompt generation, 3D propagation, and training are provided in Appendix~\ref{app:petct_sam2}.

\textbf{Encoder-based representation.} As a conventional foundation model transfer baseline, we obtain representations directly from the SAM2 image encoders (see Fig.~\ref{fig:framework}b). Using the ground-truth tumour annotation, we retain the PET and CT encoder features corresponding to tumour-containing axial slices and combine them into a 3D representation, $\mathbf{z}_{enc}^{tumour}$. This representation is provided to a CNN-based classification head,
\[
\hat{y}_{cls}=\arg\max(\mathrm{CNN}(\mathbf{z}_{enc}^{tumour})).
\]
During downstream training, the image-encoder adapters and classification head are fine-tuned, while the remaining pretrained encoder parameters remain frozen. Further details on the classification architecture and training configuration are provided in Appendix~\ref{app:encoder_baseline}.

\textbf{Object pointer representation.}
We instead construct tumour representations from the object pointers produced through segmentation (see Fig.~\ref{fig:framework}c). SAM2 produces an object pointer $\mathbf{z}_{ptr}^{i}\in\mathbb{R}^{256}$ for each processed axial slice $i$. Using the predicted segmentation, we identify the tumour-containing slices $\mathcal{S}_{tumour}^{pred}$, where $N_{tumour}^{pred}=|\mathcal{S}_{tumour}^{pred}|$ and aggregate them using mean pooling to obtain a single tumour-level representation,
\[
\mathbf{z}_{ptr}^{tumour} = \frac{1}{N_{tumour}^{pred}}\sum_{i\in\mathcal{S}_{tumour}^{pred}} \mathbf{z}_{ptr}^{i}.
\]
To probe the information already encoded in this representation $\mathbf{z}_{ptr}^{tumour}$, we use a lightweight linear prediction head,
\[
\hat{\mathbf{y}}_{cls} = \arg\max(\mathrm{softmax}\!\left(\mathbf{W}\, \mathrm{Dropout}\!\left( \mathrm{LN}\!\left(\mathbf{z}_{ptr}^{tumour}\right) \right) + \mathbf{b}\right)),
\]
where $\mathrm{LN}$ denotes layer normalisation, and $\mathbf{W}$ and $\mathbf{b}$ are the learnable parameters of the linear layer. SAM2 remains fixed after segmentation training, with downstream labels used only to train the prediction head. Downstream performance therefore probes the tumour-discriminative information already encoded by object pointers learned through segmentation. Further training and optimisation details are provided in Appendix~\ref{app:obj_ptr_seg}.

\subsection{Task-Guided Object Pointer Representations}

Segmentation-only object pointers are not explicitly optimised for downstream tasks. We therefore investigate whether directly adapting the object-centric representations to a downstream task can improve their predictive information while preserving the segmentation objective. To this end, we introduce an auxiliary classification objective directly on the aggregated tumour-level representation $\mathbf{z}_{ptr}^{tumour}$, using the same lightweight prediction head as in the segmentation-only setting (Sec.~\ref{subsec:tumour_cls}). The segmentation and classification objectives are jointly optimised,
\[
\mathcal{L}_{total} = \mathcal{L}_{seg}^{SAM} + \lambda_{\mathrm{cls}}\mathcal{L}_{cls},
\]
where $\mathcal{L}_{seg}^{SAM}$ comprises the SAM2 segmentation loss, $\mathcal{L}_{cls}$ the downstream classification loss, and $\lambda_{\mathrm{cls}}$ controls their contribution. Unlike the segmentation-only setting, gradients from $\mathcal{L}_{cls}$ are propagated through the object pointer pathway to the trainable SAM2 components, allowing downstream supervision to directly shape the learned representation while segmentation is jointly maintained.

We therefore consider two complementary settings: segmentation-only object pointers, where downstream labels are used only to train a linear prediction head over a fixed representation, and task-guided object pointers, where downstream supervision additionally shapes the object pointer representation itself. Further training and optimisation details are provided in Appendix~\ref{app:obj_ptr_aux}.

\section{Experiments}

\subsection{Experimental Setup}

\textbf{Datasets and downstream tasks.} We evaluate the learned tumour representations on two public FDG-PET/CT datasets with distinct downstream prediction tasks. Both datasets provide paired PET/CT imaging and tumour annotations for segmentation, together with information used to define the downstream tasks.

\textit{FDG-PET-CT-Lesions.} The FDG-PET-CT-Lesions dataset~\citep{https://doi.org/10.7937/gkr0-xv29} contains whole-body PET/CT imaging with manually annotated tumour lesions. From the full cohort of 900 subjects, we retain 489 subjects with melanoma, lymphoma, or lung cancer, yielding a relatively balanced three-class cancer-type classification task (see Appendix~\ref{app:dataset}). For subjects with repeated imaging, only the first imaging session is retained, so that each subject occurs once in the cohort. We use this as our primary downstream task.

\textit{HECKTOR 2022.} The HECKTOR 2022 dataset~\citep{andrearczyk2023hecktor} contains FDG-PET/CT imaging, primary tumour and lymph-node annotations, and clinical outcome information for patients with head and neck cancer collected across seven centres. 
We define a binary two-year recurrence prediction task using the primary tumour imaging and available recurrence and follow-up information. Subjects with recurrence within two years are assigned to the positive class; subjects without recurrence and with at least two years of follow-up are assigned to the negative class. Subjects with insufficient follow-up to determine their 2-year recurrence status are excluded. After cohort filtering, 427 subjects remain, including 67 positive cases (see Appendix~\ref{app:dataset}). This provides a more challenging evaluation setting due to the substantial class imbalance and the dependence of recurrence on factors beyond imaging alone. Although additional clinical variables are available, including smoking history and other patient characteristics, we use only PET/CT imaging to evaluate the information encoded by the learned imaging representations.

\textbf{Preprocessing and cohort construction.} We apply the same preprocessing pipeline to both datasets. To obtain a single tumour representation per subject, we construct unifocal cohorts by retaining the annotated primary tumour for HECKTOR 2022, and the largest connected tumour region for FDG-PET-CT-Lesions. PET intensities are represented as standardised uptake values (SUV), and PET/CT volumes are resampled to a common voxel spacing of $3\times2\times2$ mm. PET and CT intensities are clipped and globally normalised using statistics estimated from the training data. Axial slices are resized to $128\times128$ before being passed to the model. Further preprocessing and augmentation details are provided in Appendix~\ref{app:pre_aug}.

\textbf{Training and evaluation protocol.} All dataset splits are performed at the subject level and kept fixed across representation approaches. We use five-fold cross-validation, with $80\%$ of the subjects used for model development and $20\%$ held out for testing in each fold. Within the development set, $20\%$ is reserved for validation and used for model selection and hyperparameter tuning.

All SAM2-based approaches are first fine-tuned for tumour segmentation for 20 epochs. The encoder-based and task-guided approaches are initialised from the corresponding segmentation-trained checkpoint (epoch 20), and subsequently trained for the downstream task for up to 20 additional epochs. For the segmentation-only object pointers, SAM2 remains fixed after segmentation training, with the best validation checkpoint used to extract object pointers, while downstream labels are used only to optimise the lightweight prediction head. All downstream classification objectives use class-weighted cross-entropy. Further details on training, model selection, and hyperparameter tuning are provided in Appendix~\ref{app:hyperparam}.

Area under the receiver operating characteristic curve (AUC) is the primary evaluation metric, with balanced accuracy (B.Acc.) and F1 score (F1) additionally reported. Segmentation performance is evaluated using Dice similarity coefficient (Dice). Further details on evaluation metrics are provided in Appendix~\ref{app:eval_metrics}. 

Object pointer representations are constructed using predicted segmentation to identify tumour-containing slices, as described in Sec.~\ref{subsec:tumour_cls}. When segmentation fails to identify a tumour, the ground-truth annotation is used only to identify the corresponding slice indices. This ensures that the same subjects are evaluated across approaches, allowing downstream comparisons to focus on representation quality rather than differences in evaluation cohorts caused by segmentation failures.

\textbf{Baselines.} We compare object pointer representations against both learned and handcrafted tumour representations. As a conventional foundation model transfer baseline, we use the encoder-based approach described in Sec.~\ref{subsec:tumour_cls}, where the SAM2 image-encoder features are provided to a CNN prediction head for downstream classification (\textit{Encoder + CNN}). 
As a handcrafted baseline, we extract radiomics features from the ground-truth tumour masks (\textit{Radiomics})~\citep{lambin2012radiomics} and provide them to a lightweight linear prediction head analogous to that used for object pointers. 
Further details on the radiomics baseline are provided in Appendix~\ref{app:radiomics}. 

We evaluate both segmentation-only object pointers (\textit{ObjPtr (Seg.)}) and task-guided object pointers (\textit{ObjPtr (Task)}) as described in Sec.~\ref{sec:method}. To assess whether segmentation-only object pointers contain complementary information to handcrafted tumour characteristics, we additionally concatenate the object pointer and radiomics features into a combined representation (\textit{ObjPtr (Seg.) + Rad.}). 

\subsection{Experimental Results}
\begin{table}[h!]
\caption{Downstream prediction performance across representations and datasets. Results are reported as mean $\pm$ standard deviation over five-fold cross-validation. The highest mean for each metric is highlighted in \textbf{bold}.}
\label{tab:main_results}
\begin{center}
\scriptsize
\begin{tabular}{lccc|ccc}
\toprule
\multirow{2}{*}{\textbf{Method}}
& \multicolumn{3}{c|}{\textbf{FDG-PET-CT-Lesions}}
& \multicolumn{3}{c}{\textbf{HECKTOR 2022}} \\
\cmidrule(lr){2-4} \cmidrule(l){5-7}
& \textbf{AUC}$\uparrow$
& \textbf{B.Acc.}$\uparrow$
& \textbf{F1}$\uparrow$
& \textbf{AUC}$\uparrow$
& \textbf{B.Acc.}$\uparrow$
& \textbf{F1}$\uparrow$ \\
\midrule

Radiomics
& 0.833 $\pm$ 0.028
& 0.656 $\pm$ 0.046
& 0.645 $\pm$ 0.050
& 0.518 $\pm$ 0.060
& 0.517 $\pm$ 0.027
& 0.514 $\pm$ 0.029 \\

Encoder + CNN
& 0.836 $\pm$ 0.020
& 0.650 $\pm$ 0.060
& 0.635 $\pm$ 0.075
& 0.504 $\pm$ 0.106
& 0.500 $\pm$ 0.000
& 0.457 $\pm$ 0.008 \\

\midrule

ObjPtr (Seg.)
& 0.781 $\pm$ 0.057
& 0.611 $\pm$ 0.075
& 0.604 $\pm$ 0.072
& \textbf{0.631 $\pm$ 0.031}
& \textbf{0.574 $\pm$ 0.049}
& \textbf{0.542 $\pm$ 0.034} \\

ObjPtr (Task)
& \textbf{0.851 $\pm$ 0.017}
& \textbf{0.684 $\pm$ 0.024}
& \textbf{0.682 $\pm$ 0.023}
& 0.584 $\pm$ 0.055
& 0.500 $\pm$ 0.000
& 0.457 $\pm$ 0.008 \\

ObjPtr (Seg.) + Rad.
& 0.849 $\pm$ 0.019
& 0.674 $\pm$ 0.033
& 0.666 $\pm$ 0.039
& 0.549 $\pm$ 0.077
& 0.572 $\pm$ 0.066
& 0.511 $\pm$ 0.051 \\

\bottomrule
\end{tabular}
\end{center}
\end{table}

\textbf{Performance in downstream tumour prediction.} Table~\ref{tab:main_results} compares object pointer representations with radiomics and conventional encoder-based transfer. On FDG-PET-CT-Lesions, segmentation-only object pointers achieve a mean AUC of $0.781$. Although below both radiomics ($0.833$) and the encoder baseline ($0.836$), this demonstrates that object pointers learned without downstream supervision shaping the representation already encode information predictive of cancer type. 
Combining segmentation-only object pointers with radiomics increases the AUC to $0.849$, exceeding either representation individually and indicating that they contain complementary predictive information for this task. Task-guided supervision further increases performance to a mean AUC of $0.851$ with corresponding improvements in balanced accuracy and F1, while largely preserving segmentation performance (Appendix Table~\ref{tab:seg_results}). The UMAP visualisations similarly show class-dependent organisation in segmentation-only pointers and increased class separation following task-guided training (Appendix Fig.~\ref{fig:clustering}a--b).

On HECKTOR 2022,  segmentation-only object pointers achieve the highest mean performance across all metrics, including a mean AUC of $0.631$ compared with $0.518$ for radiomics and $0.504$ for the encoder baseline, both close to chance-level discrimination. In contrast to the cancer-type task, combining object pointers with radiomics reduces AUC to $0.549$, indicating that the complementary information observed on FDG-PET-CT-Lesions does not extend to recurrence prediction, where radiomics itself provides limited discriminative information. Task-guided object pointers also reduce performance relative to segmentation-only pointers, yielding an AUC of $0.584$. For both the encoder and task-guided models, the atypical balanced accuracy and F1 scores reflect collapse towards single-class predictions under the substantial class imbalance. Thus, for this much more challenging and imbalanced outcome-prediction task, downstream supervision does not improve the representation learned through segmentation.

Overall, segmentation-only object pointers encode downstream-discriminative information across both tasks despite receiving no downstream supervision during representation learning. The benefit of task-guided adaptation is task-dependent: it improves cancer-type classification, whereas the segmentation-only representation performs best for recurrence prediction. Across these settings, object pointer approaches outperform conventional encoder-based transfer and handcrafted radiomics representations.

\begin{figure}[h!]
\centering
\includegraphics[width=1.0\linewidth]{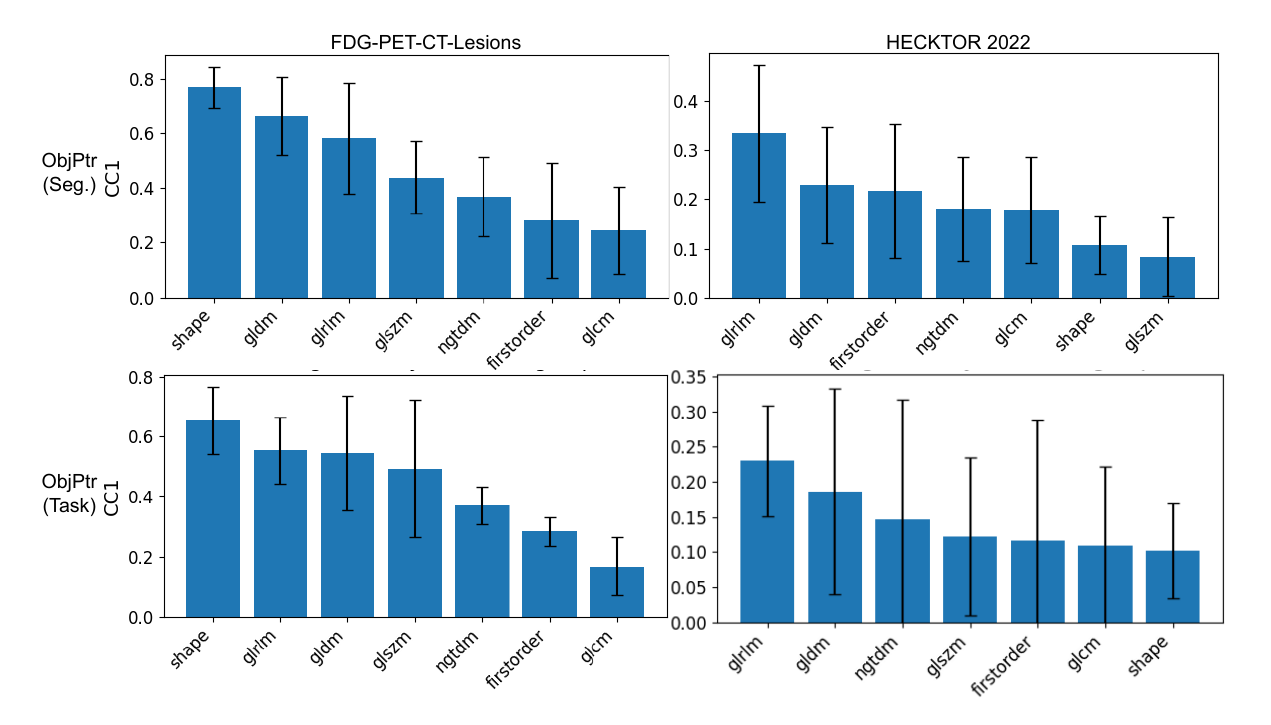}
\caption{Canonical correlation analysis (CCA) between object pointer representations and radiomics feature groups. Results are shown for segmentation-only and task-guided object pointers across FDG-PET-CT-Lesions and HECKTOR 2022. Bars show the mean canonical correlation over five-fold cross-validation, with error bars indicating standard deviation.}
\label{fig:cca}
\end{figure}

\textbf{Tumour characteristics captured by object pointers.} We next examine whether object pointer representations encode established, interpretable tumour characteristics. We use canonical correlation analysis (CCA) to quantify multivariate associations between the object pointer space and groups of radiomics features comprising tumour shape, first-order intensity, and several texture families (see Fig.~\ref{fig:cca}).

Segmentation-only object pointers (top row) show stronger association with radiomics on FDG-PET-CT-Lesions, particularly for shape- and texture-related feature groups, while considerably weaker associations are observed on HECKTOR 2022. On FDG-PET-CT-Lesions, this alignment indicates that object pointers encode information related to established radiomics tumour characteristics. However, the two representations are not redundant: combining object pointers with radiomics improves cancer-type prediction over either representation alone (see Table~\ref{tab:main_results}), providing evidence that they also contain complementary predictive information. On HECKTOR 2022, object pointers exhibit weaker associations with radiomics, whereas radiomics provide limited recurrence discrimination while segmentation-only object pointers remain more predictive. Together, these results show that segmentation-derived object pointers capture information related to established radiomics characteristics while also retaining information beyond these handcrafted features, with the relationship between the two representations varying substantially across downstream settings.

Task-guided training alters these associations, with generally weaker radiomics associations across both datasets (bottom row). Notably, on FDG-PET-CT-Lesions this occurs despite improved downstream classification, indicating that increased task performance does not require stronger associations with the predefined radiomics feature space. On HECKTOR 2022, however, weaker radiomics association is accompanied by reduced downstream performance, suggesting that decreased associations alone should not be interpreted as evidence of a more informative representation. Dimension-wise Pearson correlation analysis further identifies associations between individual object pointer dimensions and specific radiomics features (Appendix Fig.~\ref{fig:pcc_heat}).

\begin{figure}[h!]
\centering
\includegraphics[width=1.0\linewidth]{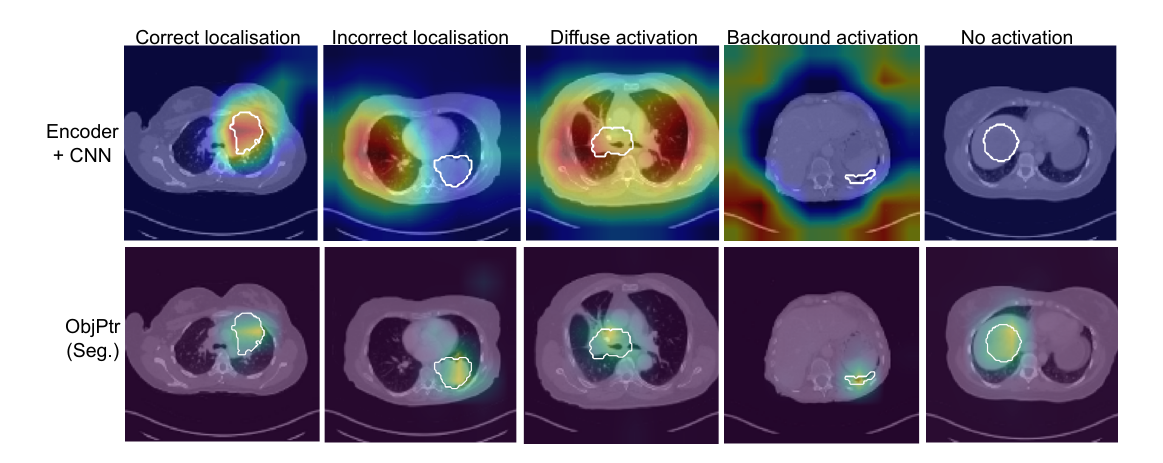}
\caption{Spatial comparison of encoder-based and object pointer representations. 
Representative FDG-PET-CT-Lesions cases spanning different spatial activation patterns of the Encoder + CNN baseline, with Grad-CAM maps shown in the top row and corresponding spatial attention maps for segmentation-only object pointers in the bottom row. White contours indicate the ground-truth tumour annotation.}
\label{fig:vis_attn}
\end{figure}

\textbf{Spatial representation analysis.} Unlike image-encoder features, object pointers emerge through object-conditioned segmentation and are therefore explicitly associated with the segmented tumour. Fig.~\ref{fig:vis_attn} qualitatively illustrates this distinction by comparing Grad-CAM~\citep{selvaraju2017grad} from the Encoder + CNN baseline with corresponding spatial attention maps from segmentation-only object pointers across representative cases from FDG-PET-CT-Lesions. While the encoder activations range from well-localised to incorrectly localised, diffuse, or absent responses, the pointer-derived maps remain concentrated around the tumour in these examples. This behaviour reflects the different origins of the two representations: encoder features are obtained before object-conditioned segmentation, whereas object pointers emerge after the model has processed the prompted tumour through the segmentation pathway. Such object conditioning may provide a useful inductive bias for downstream prediction when labelled data are limited.

\textbf{Effect of PET normalisation on representation quality.} 
We next investigate whether preprocessing choices that affect the segmentation input also influence the information retained by object pointers. We compare global PET normalisation with 3D volume-wise and 2D slice-wise z-score normalisation using segmentation-only object pointers on FDG-PET-CT-Lesions. Downstream object pointer performance decreases progressively from global to volume-wise and slice-wise normalisation, despite comparatively small differences in segmentation performance (see Appendix Table~\ref{tab:norm_results}). Volume-wise normalisation removes inter-patient intensity variation, while slice-wise normalisation removes variation across slices within each volume, potentially discarding information relevant to downstream tumour discrimination. These results show that preprocessing choices can affect representation quality even when their impact on the primary segmentation task is relatively limited. Consistent with this trend, the UMAP visualisation shows progressively less distinct class-dependent structure from global to volume-wise and slice-wise normalisation (Appendix Fig.~\ref{fig:clustering}a,c,d).

\textbf{Auxiliary loss ablation.} We assess the effect of the classification-loss weight \(\lambda_{\mathrm{cls}}\) on the task-guided representation on FDG-PET-CT-Lesions. Increasing \(\lambda_{\mathrm{cls}}\) does not meaningfully improve downstream classification performance, while segmentation performance progressively decreases (see Appendix Fig.~\ref{fig:reg_ablation}). 
This suggests that placing greater weight on the downstream objective provides no additional predictive benefit while increasingly compromising the segmentation task. We therefore use \(\lambda_{\mathrm{cls}}=1\) in the main experiments.

\section{Conclusion}
In this work, we investigated SAM2 object pointers as transferable object-level representations for downstream tumour prediction in PET/CT imaging. We show that object pointers emerging from segmentation alone already encode information predictive of downstream targets, despite receiving no downstream supervision during representation learning. The utility of further task-specific adaptation is task-dependent: task-guided object pointers improve cancer-type classification, whereas segmentation-only representations perform best for recurrence prediction.

Our analyses show that object pointers are associated with interpretable radiomics tumour characteristics and exhibit class-dependent organisation within the learned representation space. For cancer-type classification, combining segmentation-only object pointers with radiomics improves performance over either representation alone, indicating complementary predictive information. Qualitative spatial analysis additionally illustrates the tumour-centred nature of representations emerging through object-conditioned segmentation. Finally, we find that preprocessing choices can substantially affect downstream representation quality despite comparatively smaller changes in segmentation performance, highlighting that primary-task performance alone does not fully characterise the representations learned through that task.

Together, these findings demonstrate that useful transferable representations can emerge through the execution of a foundation model's primary task, rather than being confined to its encoder. More broadly, they motivate considering task-emergent internal representations as an additional source of transferable information in structured foundation models. Future work could investigate whether this behaviour generalises across foundation models and imaging domains.

\subsection*{AI use statement}
Generative AI tools were used to assist with language editing, improving clarity, and refining the presentation of the manuscript. All scientific content, methodological decisions, experiments, analyses, and conclusions were developed and verified by the authors, who take full responsibility for the final manuscript.

\subsection*{Ethics statement}
This study uses publicly available, de-identified medical imaging datasets and involves no prospective data collection or direct interaction with patients. We use the data solely for retrospective computational analysis in accordance with the terms of the respective datasets. The proposed methods are intended for research purposes and are not designed or evaluated for direct clinical decision-making. Any clinical translation would require further validation across independent and representative patient populations.

\subsection*{Reproducibility statement}
We are committed to ensuring the reproducibility of our results. We provide detailed descriptions of the datasets and preprocessing, model implementation and training, and evaluation procedures in the main text and Appendix. Additionally, all datasets used in this paper are publicly available. The source code will be made publicly available upon publication.

\bibliography{iclr2027_conference}

@inproceedings{ravi2025sam,
  title={Sam 2: Segment anything in images and videos},
  author={Ravi, Nikhila and Gabeur, Valentin and Hu, Yuan-Ting and Hu, Ronghang and Ryali, Chaitanya and Ma, Tengyu and Khedr, Haitham and R{\"a}dle, Roman and Rolland, Chloe and Gustafson, Laura and others},
  booktitle={International Conference on Learning Representations},
  volume={2025},
  pages={28085--28128},
  year={2025}
}

@article{jiang2023deep,
  title={Deep learning for medical image-based cancer diagnosis},
  author={Jiang, Xiaoyan and Hu, Zuojin and Wang, Shuihua and Zhang, Yudong},
  journal={Cancers},
  volume={15},
  number={14},
  pages={3608},
  year={2023},
  publisher={MDPI}
}

@article{gillies2016radiomics,
  title={Radiomics: images are more than pictures, they are data},
  author={Gillies, Robert J and Kinahan, Paul E and Hricak, Hedvig},
  journal={Radiology},
  volume={278},
  number={2},
  pages={563--577},
  year={2016},
  publisher={Radiological Society of North America}
}

@article{erickson2017machine,
  title={Machine learning for medical imaging},
  author={Erickson, Bradley J and Korfiatis, Panagiotis and Akkus, Zeynettin and Kline, Timothy L},
  journal={radiographics},
  volume={37},
  number={2},
  pages={505--515},
  year={2017},
  publisher={Radiological Society of North America}
}

@article{bi2019artificial,
  title={Artificial intelligence in cancer imaging: clinical challenges and applications},
  author={Bi, Wenya Linda and Hosny, Ahmed and Schabath, Matthew B and Giger, Maryellen L and Birkbak, Nicolai J and Mehrtash, Alireza and Allison, Tavis and Arnaout, Omar and Abbosh, Christopher and Dunn, Ian F and others},
  journal={CA: a cancer journal for clinicians},
  volume={69},
  number={2},
  pages={127--157},
  year={2019},
  publisher={Wiley Online Library}
}

@article{koh2022artificial,
  title={Artificial intelligence and machine learning in cancer imaging},
  author={Koh, Dow-Mu and Papanikolaou, Nickolas and Bick, Ulrich and Illing, Rowland and Kahn Jr, Charles E and Kalpathi-Cramer, Jayshree and Matos, Celso and Mart{\'\i}-Bonmat{\'\i}, Luis and Miles, Anne and Mun, Seong Ki and others},
  journal={Communications Medicine},
  volume={2},
  number={1},
  pages={133},
  year={2022},
  publisher={Nature Publishing Group UK London}
}

@article{fass2008imaging,
  title={Imaging and cancer: a review},
  author={Fass, Leonard},
  journal={Molecular oncology},
  volume={2},
  number={2},
  pages={115--152},
  year={2008},
  publisher={Wiley Online Library}
}

@article{moor2023foundation,
  title={Foundation models for generalist medical artificial intelligence},
  author={Moor, Michael and Banerjee, Oishi and Abad, Zahra Shakeri Hossein and Krumholz, Harlan M and Leskovec, Jure and Topol, Eric J and Rajpurkar, Pranav},
  journal={Nature},
  volume={616},
  number={7956},
  pages={259--265},
  year={2023},
  publisher={Nature Publishing Group UK London}
}

@article{aerts2014decoding,
  title={Decoding tumour phenotype by noninvasive imaging using a quantitative radiomics approach},
  author={Aerts, Hugo JWL and Velazquez, Emmanuel Rios and Leijenaar, Ralph TH and Parmar, Chintan and Grossmann, Patrick and Carvalho, Sara and Bussink, Johan and Monshouwer, Ren{\'e} and Haibe-Kains, Benjamin and Rietveld, Derek and others},
  journal={Nature communications},
  volume={5},
  number={1},
  pages={4006},
  year={2014},
  publisher={Nature Publishing Group UK London}
}

@article{rani2024self,
  title={Self-supervised learning for medical image analysis: a comprehensive review},
  author={Rani, Veenu and Kumar, Munish and Gupta, Aastha and Sachdeva, Monika and Mittal, Ajay and Kumar, Krishan},
  journal={Evolving Systems},
  volume={15},
  number={4},
  pages={1607--1633},
  year={2024},
  publisher={Springer}
}

@article{shurrab2022self,
  title={Self-supervised learning methods and applications in medical imaging analysis: A survey},
  author={Shurrab, Saeed and Duwairi, Rehab},
  journal={PeerJ Computer Science},
  volume={8},
  pages={e1045},
  year={2022},
  publisher={PeerJ Inc.}
}

@article{lambin2012radiomics,
  title={Radiomics: extracting more information from medical images using advanced feature analysis},
  author={Lambin, Philippe and Rios-Velazquez, Emmanuel and Leijenaar, Ralph and Carvalho, Sara and Van Stiphout, Ruud GPM and Granton, Patrick and Zegers, Catharina ML and Gillies, Robert and Boellard, Ronald and Dekker, Andr{\'e} and others},
  journal={European journal of cancer},
  volume={48},
  number={4},
  pages={441--446},
  year={2012},
  publisher={Elsevier}
}

@article{yip2016applications,
  title={Applications and limitations of radiomics},
  author={Yip, Stephen SF and Aerts, Hugo JWL},
  journal={Physics in Medicine \& Biology},
  volume={61},
  number={13},
  pages={R150--R166},
  year={2016},
  publisher={IOP Publishing}
}

@inproceedings{kirillov2023segment,
  title={Segment anything},
  author={Kirillov, Alexander and Mintun, Eric and Ravi, Nikhila and Mao, Hanzi and Rolland, Chloe and Gustafson, Laura and Xiao, Tete and Whitehead, Spencer and Berg, Alexander C and Lo, Wan-Yen and others},
  booktitle={2023 IEEE/CVF international conference on computer vision (ICCV)},
  pages={3992--4003},
  year={2023},
  organization={IEEE}
}

@article{ma2024segment,
  title={Segment anything in medical images},
  author={Ma, Jun and He, Yuting and Li, Feifei and Han, Lin and You, Chenyu and Wang, Bo},
  journal={Nature communications},
  volume={15},
  number={1},
  pages={654},
  year={2024},
  publisher={Nature Publishing Group UK London}
}

@article{xiang2025vision,
  title={A vision--language foundation model for precision oncology},
  author={Xiang, Jinxi and Wang, Xiyue and Zhang, Xiaoming and Xi, Yinghua and Eweje, Feyisope and Chen, Yijiang and Li, Yuchen and Bergstrom, Colin and Gopaulchan, Matthew and Kim, Ted and others},
  journal={Nature},
  volume={638},
  number={8051},
  pages={769--778},
  year={2025},
  publisher={Nature Publishing Group UK London}
}

@article{zhou2023foundation,
  title={A foundation model for generalizable disease detection from retinal images},
  author={Zhou, Yukun and Chia, Mark A and Wagner, Siegfried K and Ayhan, Murat S and Williamson, Dominic J and Struyven, Robbert R and Liu, Timing and Xu, Moucheng and Lozano, Mateo G and Woodward-Court, Peter and others},
  journal={Nature},
  volume={622},
  number={7981},
  pages={156--163},
  year={2023},
  publisher={Nature Publishing Group UK London}
}

@article{huang2025systematic,
  title={A systematic review and implementation guidelines of multimodal foundation models in medical imaging},
  author={Huang, Shih-Cheng and Jensen, Malte and Yeung-Levy, Serena and Lungren, Matthew P and Poon, Hoifung and Chaudhari, Akshay S},
  journal={Research Square},
  pages={rs--3},
  year={2025}
}

@article{pai2024foundation,
  title={Foundation model for cancer imaging biomarkers},
  author={Pai, Suraj and Bontempi, Dennis and Hadzic, Ibrahim and Prudente, Vasco and Soka{\v{c}}, Mateo and Chaunzwa, Tafadzwa L and Bernatz, Simon and Hosny, Ahmed and Mak, Raymond H and Birkbak, Nicolai J and others},
  journal={Nature machine intelligence},
  volume={6},
  number={3},
  pages={354--367},
  year={2024},
  publisher={Nature Publishing Group UK London}
}

@inproceedings{gu2025adapting,
  title={Adapting a segmentation foundation model for medical image classification},
  author={Gu, Pengfei and Tang, Haoteng and Ebeid, Islam A and Nunez, Jose A and Vazquez, Fabian and Adame, Diego and Zhan, Marcus and Li, Huimin and Fu, Bin and Chen, Danny Z},
  booktitle={2025 IEEE 38th International Symposium on Computer-Based Medical Systems (CBMS)},
  pages={167--172},
  year={2025},
  organization={IEEE}
}

@article{jacob2025towards,
  title={Towards a cardiovascular magnetic resonance foundation model for multi-task cardiac image analysis},
  author={Jacob, Athira J and Borgohain, Indraneel and Chitiboi, Teodora and Sharma, Puneet and Comaniciu, Dorin and Rueckert, Daniel},
  journal={Journal of Cardiovascular Magnetic Resonance},
  volume={27},
  number={2},
  pages={101967},
  year={2025},
  publisher={Elsevier}
}

@inproceedings{wu2024voco,
  title={Voco: A simple-yet-effective volume contrastive learning framework for 3d medical image analysis},
  author={Wu, Linshan and Zhuang, Jiaxin and Chen, Hao},
  booktitle={2024 IEEE/CVF Conference on Computer Vision and Pattern Recognition (CVPR)},
  pages={22873--22882},
  year={2024},
  organization={IEEE}
}

@misc{https://doi.org/10.7937/gkr0-xv29,
  doi = {10.7937/GKR0-XV29},
  url = {https://www.cancerimagingarchive.net/collection/fdg-pet-ct-lesions/},
  author = {Gatidis, Sergios and Kuestner, Thomas},
  title = {A whole-body FDG-PET/CT dataset with manually annotated tumor lesions (FDG-PET-CT-Lesions)},
  publisher = {The Cancer Imaging Archive},
  year = {2022},
  copyright = {NIH Controlled Data Access Policy}
}

@inproceedings{andrearczyk2023hecktor,
  title={Overview of the HECKTOR Challenge at MICCAI 2022: Automatic Head and Neck Tumor Segmentation and Outcome Prediction in PET/CT},
  author={Andrearczyk, Vincent and Oreiller, Valentin and Buvat, Irene and Hatt, Martin and Depeursinge, Adrien},
  booktitle={Head and Neck Tumor Segmentation and Outcome Prediction},
  pages={1--25},
  year={2023},
  publisher={Springer, Cham},
  doi={10.1007/978-3-031-27420-6_1}
}

@article{wang2023seggpt,
  title={Seggpt: Segmenting everything in context},
  author={Wang, Xinlong and Zhang, Xiaosong and Cao, Yue and Wang, Wen and Shen, Chunhua and Huang, Tiejun},
  journal={arXiv preprint arXiv:2304.03284},
  year={2023}
}

@article{zou2023segment,
  title={Segment everything everywhere all at once},
  author={Zou, Xueyan and Yang, Jianwei and Zhang, Hao and Li, Feng and Li, Linjie and Wang, Jianfeng and Wang, Lijuan and Gao, Jianfeng and Lee, Yong Jae},
  journal={Advances in neural information processing systems},
  volume={36},
  pages={19769--19782},
  year={2023}
}

@article{avanzo2020machine,
  title={Machine and deep learning methods for radiomics},
  author={Avanzo, Michele and Wei, Lise and Stancanello, Joseph and Vallieres, Martin and Rao, Arvind and Morin, Olivier and Mattonen, Sarah A and El Naqa, Issam},
  journal={Medical physics},
  volume={47},
  number={5},
  pages={e185--e202},
  year={2020},
  publisher={Wiley Online Library}
}

@article{liu2019applications,
  title={The applications of radiomics in precision diagnosis and treatment of oncology: opportunities and challenges},
  author={Liu, Zhenyu and Wang, Shuo and Dong, Di and Wei, Jingwei and Fang, Cheng and Zhou, Xuezhi and Sun, Kai and Li, Longfei and Li, Bo and Wang, Meiyun and others},
  journal={Theranostics},
  volume={9},
  number={5},
  pages={1303},
  year={2019}
}

@article{yang2025foundation,
  title={A foundation model for generalizable cancer diagnosis and survival prediction from histopathological images},
  author={Yang, Zhaochang and Wei, Ting and Liang, Ying and Yuan, Xin and Gao, Ruitian and Xia, Yujia and Zhou, Jie and Zhang, Yue and Yu, Zhangsheng},
  journal={Nature Communications},
  volume={16},
  number={1},
  pages={2366},
  year={2025},
  publisher={Nature Publishing Group UK London}
}

@article{ryu2025vision,
  title={Vision-language foundation models for medical imaging: a review of current practices and innovations},
  author={Ryu, Ji Seung and Kang, Hyunyoung and Chu, Yuseong and Yang, Sejung},
  journal={Biomedical Engineering Letters},
  volume={15},
  number={5},
  pages={809--830},
  year={2025},
  publisher={Springer}
}

@inproceedings{huang2024comparative,
  title={Comparative analysis of imagenet pre-trained deep learning models and dinov2 in medical imaging classification},
  author={Huang, Yuning and Zou, Jingchen and Meng, Lanxi and Yue, Xin and Zhao, Qing and Li, Jianqiang and Song, Changwei and Jimenez, Gabriel and Li, Shaowu and Fu, Guanghui},
  booktitle={2024 IEEE 48th Annual Computers, Software, and Applications Conference (COMPSAC)},
  pages={297--305},
  year={2024},
  organization={IEEE}
}

@inproceedings{caron2021emerging,
  title={Emerging properties in self-supervised vision transformers},
  author={Caron, Mathilde and Touvron, Hugo and Misra, Ishan and J{\'e}gou, Herv{\'e} and Mairal, Julien and Bojanowski, Piotr and Joulin, Armand},
  booktitle={2021 IEEE/CVF international conference on computer vision (ICCV)},
  pages={9630--9640},
  year={2021},
  organization={IEEE}
}

@article{oquab2023dinov2,
  title={Dinov2: Learning robust visual features without supervision},
  author={Oquab, Maxime and Darcet, Timoth{\'e}e and Moutakanni, Th{\'e}o and Vo, Huy and Szafraniec, Marc and Khalidov, Vasil and Fernandez, Pierre and Haziza, Daniel and Massa, Francisco and El-Nouby, Alaaeldin and others},
  journal={arXiv preprint arXiv:2304.07193},
  year={2023}
}

@article{baharoon2023evaluating,
  title={Evaluating general purpose vision foundation models for medical image analysis: An experimental study of dinov2 on radiology benchmarks},
  author={Baharoon, Mohammed and Qureshi, Waseem and Ouyang, Jiahong and Xu, Yanwu and Aljouie, Abdulrhman and Peng, Wei},
  journal={arXiv preprint arXiv:2312.02366},
  year={2023}
}

@article{prokop2023deep,
  title={Deep metric learning for few-shot X-ray image classification},
  author={Prokop, Jakub and Tordera, Javier Montalt and Jaworek-Korjakowska, Joanna and Mohammadi, Sadegh},
  journal={MedRxiv},
  pages={2023--08},
  year={2023},
  publisher={Cold Spring Harbor Laboratory Press}
}

@inproceedings{huix2024natural,
  title={Are natural domain foundation models useful for medical image classification?},
  author={Huix, Joana Pal{\'e}s and Ganeshan, Adithya Raju and Haslum, Johan Fredin and S{\"o}derberg, Magnus and Matsoukas, Christos and Smith, Kevin},
  booktitle={2024 IEEE/CVF Winter Conference on Applications of Computer Vision (WACV)},
  pages={7619--7628},
  year={2024},
  organization={IEEE}
}

@article{paschali2025foundation,
  title={Foundation models in radiology: what, how, why, and why not},
  author={Paschali, Magdalini and Chen, Zhihong and Blankemeier, Louis and Varma, Maya and Youssef, Alaa and Bluethgen, Christian and Langlotz, Curtis and Gatidis, Sergios and Chaudhari, Akshay},
  journal={Radiology},
  volume={314},
  number={2},
  pages={e240597},
  year={2025},
  publisher={Radiological Society of North America}
}

@article{sun2024medical,
  title={Medical image analysis using improved SAM-Med2D: segmentation and classification perspectives},
  author={Sun, Jiakang and Chen, Ke and He, Zhiyi and Ren, Siyuan and He, Xinyang and Liu, Xu and Peng, Cheng},
  journal={BMC Medical Imaging},
  volume={24},
  number={1},
  pages={241},
  year={2024},
  publisher={Springer}
}

@article{binzagr2026foundation,
  title={Foundation-Model-Driven Skin Lesion Segmentation and Classification Using SAM-Adapters and Vision Transformers},
  author={Binzagr, Faisal and Hariri, Majed},
  journal={Diagnostics},
  volume={16},
  number={3},
  pages={468},
  year={2026},
  publisher={MDPI}
}

@article{wang2024sam,
  title={SAM-IE: SAM-based image enhancement for facilitating medical image diagnosis with segmentation foundation model},
  author={Wang, Changyan and Chen, Haobo and Zhou, Xin and Wang, Meng and Zhang, Qi},
  journal={Expert Systems with Applications},
  volume={249},
  pages={123795},
  year={2024},
  publisher={Elsevier}
}

@article{rubinstein2025we,
  title={Are we done with object-centric learning?},
  author={Rubinstein, Alexander and Prabhu, Ameya and Bethge, Matthias and Oh, Seong Joon},
  journal={arXiv preprint arXiv:2504.07092},
  year={2025}
}

@article{chen2024sam2,
  title={Sam2-adapter: Evaluating \& adapting segment anything 2 in downstream tasks: Camouflage, shadow, medical image segmentation, and more},
  author={Chen, Tianrun and Lu, Ankang and Zhu, Lanyun and Ding, Chaotao and Yu, Chunan and Ji, Deyi and Li, Zejian and Sun, Lingyun and Mao, Papa and Zang, Ying},
  journal={arXiv preprint arXiv:2408.04579},
  year={2024}
}

@inproceedings{ryali2023hiera,
  title={Hiera: A hierarchical vision transformer without the bells-and-whistles},
  author={Ryali, Chaitanya and Hu, Yuan-Ting and Bolya, Daniel and Wei, Chen and Fan, Haoqi and Huang, Po-Yao and Aggarwal, Vaibhav and Chowdhury, Arkabandhu and Poursaeed, Omid and Hoffman, Judy and others},
  booktitle={International conference on machine learning},
  pages={29441--29454},
  year={2023},
  organization={PMLR}
}

@article{dong2026segment,
  title={Segment anything model 2: an application to 2D and 3D medical images},
  author={Dong, Haoyu and Gu, Hanxue and Chen, Yaqian and Yang, Jichen and Chen, Yuwen and Mazurowski, Maciej A},
  journal={IEEE Transactions on Biomedical Engineering},
  year={2026},
  publisher={IEEE}
}

@article{cardoso2022monai,
  title={Monai: An open-source framework for deep learning in healthcare},
  author={Cardoso, M Jorge and Li, Wenqi and Brown, Richard and Ma, Nic and Kerfoot, Eric and Wang, Yiheng and Murrey, Benjamin and Myronenko, Andriy and Zhao, Can and Yang, Dong and others},
  journal={arXiv preprint arXiv:2211.02701},
  year={2022}
}

@article{van2017computational,
  title={Computational radiomics system to decode the radiographic phenotype},
  author={Van Griethuysen, Joost JM and Fedorov, Andriy and Parmar, Chintan and Hosny, Ahmed and Aucoin, Nicole and Narayan, Vivek and Beets-Tan, Regina GH and Fillion-Robin, Jean-Christophe and Pieper, Steve and Aerts, Hugo JWL},
  journal={Cancer research},
  volume={77},
  number={21},
  pages={e104--e107},
  year={2017},
  publisher={American Association for Cancer Research}
}

@article{mcinnes2018umap,
  title={Umap: Uniform manifold approximation and projection for dimension reduction},
  author={McInnes, Leland and Healy, John and Melville, James},
  journal={arXiv preprint arXiv:1802.03426},
  year={2018}
}

@article{wu2025rethinking,
  title={Rethinking foundation models for medical image classification through a benchmark study on medmnist},
  author={Wu, Fuping and Papiez, Bartlomiej W},
  journal={arXiv preprint arXiv:2501.14685},
  year={2025}
}

@inproceedings{selvaraju2017grad,
  title={Grad-cam: Visual explanations from deep networks via gradient-based localization},
  author={Selvaraju, Ramprasaath R and Cogswell, Michael and Das, Abhishek and Vedantam, Ramakrishna and Parikh, Devi and Batra, Dhruv},
  booktitle={Proceedings of the IEEE international conference on computer vision},
  pages={618--626},
  year={2017}
}
\bibliographystyle{iclr2027_conference}

\appendix
\section{Appendix}

\subsection{SAM2 PET/CT Segmentation}
\label{app:petct_sam2}

\subsubsection{Architecture and Adaptation}

Our model is initialised from the SAM2-Base+ checkpoint, which uses a four-stage hierarchical vision transformer (Hiera)~\citep{ryali2023hiera} and comprises 80.8M parameters. The PET and CT image encoders are independently initialised from the pretrained SAM2 image encoder, with the original encoder parameters kept frozen throughout training. Modality-specific trainable adapters are inserted at the 12th (middle) and 24th (final) Hiera blocks for parameter-efficient adaptation. Each adapter consists of a down-projection from the original feature dimension to a dimension of 64, followed by a GeLU activation, an up-projection back to the original feature dimension, and a second GeLU activation. The resulting PET and CT embeddings are fused after the image encoder using the learnable weighting parameter $\alpha$ described in the main text, initialised to $0.5$. The prompt encoder, mask decoder, image-encoder adapters, and modality-fusion parameter are optimised during segmentation training, while the remaining pretrained SAM2 parameters are kept frozen.

\subsubsection{Prompt Generation}

SAM2 supports point-, box-, and mask-based prompts. We use a bounding-box prompt as it provides explicit localisation of the tumour while requiring only a simple spatial annotation. A single 2D bounding box is generated from the ground-truth tumour annotation on the central axial tumour slice, which typically contains a large and clearly visible cross-section of the tumour. The bounding box is randomly perturbed by independently expanding each boundary by up to 10\% of the corresponding box dimension. Perturbations are resampled at each training epoch and are also applied during validation and testing, simulating variability in manual box placement and preventing the model from relying on precisely defined prompts. 

\subsubsection{3D Propagation}

Starting from the centrally prompted axial slice, segmentation is propagated through the 3D volume. Propagation is first performed from the central slice towards one end of the volume (backward). The model is then returned to the prompted slice and the memory is reset to prevent information from the backward propagation influencing predictions in the opposite direction~\citep{dong2026segment}. Propagation is subsequently performed towards the other end of the volume (forward), allowing the tumour to be segmented on both sides of the initial prompt. No additional prompts are provided to subsequent slices.

\subsubsection{SAM2 loss}

We use the standard SAM2 segmentation loss components with adapted relative weighting. The segmentation objective is defined as:
\[
L_{seg}^{SAM} = 0.4 \cdot L_{focal} + 0.4 \cdot L_{dice} + 0.1 \cdot L_{IoU} + 0.1 \cdot L_{class},
\]
Here, $\mathcal{L}_{focal}$ is a sigmoid focal loss and $\mathcal{L}_{dice}$ is a Dice loss, which together supervise mask prediction. $\mathcal{L}_{IoU}$ supervises the predicted mask-quality score, while $\mathcal{L}_{class}$ supervises object presence. Lower weights are assigned to the IoU and object-presence losses since our setting considers a single object of interest and uses single-mask prediction, reducing the importance of object-presence and mask-selection supervision, respectively.

\subsubsection{Training Configuration}
\label{subsec:sam2_training}

Training largely follows the SAM2 protocol with minor adjustments. We use stochastic gradient descent (SGD) with a momentum of $0.99$ and a polynomial learning-rate scheduler with power $0.9$. The initial learning rate is set to $0.01$ with a weight decay of $3\times10^{-5}$. Gradients are clipped to a maximum norm of $1.0$ for training stability. Models are trained for $20$ epochs with a batch size of $8$.

\subsection{Encoder-Based Classification Baseline}
\label{app:encoder_baseline}

\subsubsection{CNN Classification Head}

The PET and CT image-encoder embeddings are concatenated along the feature dimension, and the embeddings corresponding to tumour-containing axial slices $\mathcal{S}_{tumour}^{GT}$ are retained using the ground-truth tumour annotation,
\[
\mathbf{z}_{enc}^{tumour}
=
\operatorname{Concat}\left(
\mathbf{z}_{PET}[\mathcal{S}_{tumour}^{GT}],
\mathbf{z}_{CT}[\mathcal{S}_{tumour}^{GT}]
\right).
\]
The resulting 3D feature volume is passed to a CNN classification head consisting of two $3\times3\times3$ convolutional layers with 128 and 64 channels, respectively. Each convolutional layer is followed by Group Normalisation, ReLU activation, and dropout ($p=0.2$). Adaptive average pooling reduces the feature volume to a single $1\times1\times1$ representation, followed by fully connected layers with 32 and $C$ units, where $C$ is the number of downstream classes. ReLU activation and dropout are applied between the fully connected layers, and a softmax operation produces the final class probabilities. For cases with fewer than three tumour-containing slices, the slice dimension is zero-padded to a minimum depth of three to accommodate the $3\times3\times3$ convolutional kernels.

\subsubsection{Training Configuration}

The encoder-based classification model is initialised from the corresponding segmentation-trained checkpoint. During downstream training, the image-encoder adapters and CNN classification head are optimised, while the remaining pretrained SAM2 parameters remain frozen. The CNN classification head is optimised using AdamW with a learning rate of $1\times10^{-3}$, weight decay of $1\times10^{-2}$, and dropout $p=0.2$.  Classification is trained using class-weighted cross-entropy loss with a batch size of $8$, and gradients are clipped to a maximum norm of $1.0$. The optimal training epoch is selected based on validation performance, as detailed in Sec.~\ref{app:hyperparam}.

\subsection{object pointer Classification}
\label{app:obj_ptr}

\subsubsection{Segmentation-only Classification}
\label{app:obj_ptr_seg}

For the segmentation-only setting, the object pointer representations obtained from the segmentation-trained model are kept fixed, and only the downstream prediction head is trained. The classifier is optimised using AdamW for up to 100 epochs. Learning rate, weight decay, dropout, and batch size are selected using the validation set, as detailed in Sec.~\ref{app:hyperparam}

\subsubsection{Task-guided Training}
\label{app:obj_ptr_aux}

For task-guided object pointers, the prediction head is trained jointly with the segmentation model. The prediction head is optimised using AdamW with a learning rate of $1\times10^{-3}$, weight decay of $1\times10^{-2}$, and dropout $p=0.2$. The segmentation model is optimised as described in Sec.~\ref{subsec:sam2_training}. Training uses a batch size of $8$ and gradients are clipped to a maximum norm of $1.0$. The optimal training epoch is selected based on validation performance, as detailed in Sec.~\ref{app:hyperparam}.

\subsection{Dataset and Cohort Details}
\label{app:dataset}

\subsubsection{FDG-PET-CT-Lesions}

The final FDG-PET-CT-Lesions cohort consists of 489 subjects, including 177 melanoma, 168 lung cancer, and 144 lymphoma cases, resulting in a relatively balanced distribution across the three cancer types.

\subsubsection{HECKTOR 2022}

Of the 524 subjects in HECKTOR 2022, recurrence information is available for 488. Subjects without recurrence and with less than two years of follow-up are excluded ($n=48$), as their 2-year recurrence status cannot be determined, leaving 440 subjects. A further 13 subjects without an identified primary tumour are excluded, resulting in a final cohort of 427 subjects. Of these, 67 experience recurrence within two years and 360 have no recurrence with at least two years of follow-up.

\subsection{Image Preprocessing and Augmentation}
\label{app:pre_aug}

\subsubsection{Image Preprocessing}

PET intensities are represented as standardised uptake values (SUV). PET/CT volumes and corresponding tumour annotations are resampled to a common voxel spacing of $3\times2\times2$ mm. Volumes are cropped or padded in-plane to $200\times200$ voxels and limited to a maximum depth of 500 axial slices, while ensuring that the tumour region remains within the retained volume. PET and CT intensities are clipped to the $0.5$th--$99.5$th percentiles and globally z-score normalised using statistics estimated from the training data within each cross-validation fold. Axial slices are resized to $128\times128$ before being passed to the model, while the out-of-plane dimension was retained to accommodate variable numbers of slices across scans. Following resizing, two very small tumour annotations in FDG-PET-CT-Lesions contained no remaining foreground voxels and were therefore ignored during training and evaluation. No such cases occurred in HECKTOR 2022. As the SAM2 image encoder expects three-channel input, each single-channel modality is replicated across the channel dimension, resulting in \([\mathrm{PET},\mathrm{PET},\mathrm{PET}]\) and \([\mathrm{CT},\mathrm{CT},\mathrm{CT}]\).

\subsubsection{Data Augmentation}

Data augmentation is implemented using MONAI (v1.3.2)~\citep{cardoso2022monai} and is primarily spatial to preserve potentially informative PET/CT intensity characteristics. Spatial augmentations are applied jointly to PET, CT, and the tumour annotation to maintain their spatial correspondence. These include random flipping with probability $0.5$, random rotation with probability $0.3$ and a maximum angle of $30^\circ$, and random zooming with probability $0.25$ over a range of $0.8$--$1.2$. Gaussian noise with zero mean and standard deviation $0.1$ is additionally applied independently to each imaging modality with probability $0.1$.

\subsection{Radiomics Baseline}
\label{app:radiomics}

\subsubsection{Feature Extraction}

Radiomics features are extracted from the ground-truth tumour volumes using PyRadiomics (v3.0.1)~\citep{van2017computational}. PET and CT volumes undergo the same preprocessing described in Sec.~\ref{app:pre_aug} prior to feature extraction. The voxel spacing supplied to PyRadiomics is adjusted to account for the spatial resizing, such that physical voxel dimensions are retained during feature extraction. Features are computed from both PET and CT images and include first-order intensity, shape, GLCM, GLRLM, GLSZM, GLDM, and NGTDM features. For each modality, 107 features were computed and concatenated, with duplicate entries removed, resulting in a 200-dimensional feature vector.

\subsubsection{Downstream Classification}

The extracted radiomics features are provided to the same lightweight prediction head used for the object pointer representations. The classifier is trained for up to 100 epochs using AdamW, with learning rate, weight decay, dropout, and batch size selected on the validation set as detailed in Sec.~\ref{app:hyperparam}.

\subsection{Model Selection and Hyperparameter Tuning}
\label{app:hyperparam}

All model selection and hyperparameter tuning are performed independently within each cross-validation fold using the corresponding validation set. Held-out test subjects are used only for final evaluation.

\subsubsection{Segmentation Model}

SAM2 is trained for tumour segmentation for 20 epochs. The checkpoint with the highest validation Dice among the final five epochs is selected for segmentation evaluation.

\subsubsection{Encoder + CNN}
The encoder-based model is initialised from the final segmentation checkpoint at epoch 20 and trained for up to 20 additional epochs. Here, the segmentation checkpoint serves only as an initialisation before end-to-end downstream fine-tuning, so the final checkpoint from segmentation training is used rather than selecting an intermediate checkpoint based on segmentation performance. The checkpoint achieving the highest validation AUC among the final five epochs is selected for final evaluation.

\subsubsection{Segmentation-only object pointers}
Object pointer representations are extracted using the segmentation checkpoint with the highest validation Dice among the final five epochs. Since these representations remain fixed during downstream training, the validation-selected segmentation checkpoint is used to provide the best segmentation-derived representation before training the lightweight prediction head. The downstream prediction head is trained for up to 100 epochs. Hyperparameters are selected using a random search over 100 unique configurations. The learning rate is sampled from $\{10^{-4}, 5\times10^{-4}, 10^{-3}, 5\times10^{-3}, 10^{-2}\}$, batch size from $\{8,16,32,64\}$, weight decay from $\{10^{-4}, 5\times10^{-4}, 10^{-3}, 5\times10^{-3}, 10^{-2}\}$, and dropout from $\{0,0.1,0.2,0.3,0.4,0.5\}$. The configuration and checkpoint achieving the highest validation AUC are selected independently within each fold.

\subsubsection{Task-guided object pointers}
The task-guided model is initialised from the final segmentation checkpoint at epoch 20 and trained for up to 20 additional epochs. As in the encoder-based approach, this checkpoint serves only as an initialisation before end-to-end downstream fine-tuning. The checkpoint achieving the highest validation AUC among the final five epochs is selected for final evaluation. The auxiliary classification-loss weight is set to \(\lambda_{\mathrm{cls}}=1\) based on the validation ablation reported in Appendix Fig.~\ref{fig:reg_ablation}.

\subsubsection{Radiomics and combined representation}
The radiomics classifier and the combined radiomics and segmentation-only object pointer classifiers follow the same hyperparameter-tuning and model-selection protocol as the segmentation-only object pointer prediction head, with learning rate, weight decay, dropout, and batch size selected using validation AUC.

\subsection{Evaluation Metrics}
\label{app:eval_metrics}

For downstream classification, we report area under the receiver operating characteristic curve (AUC), balanced accuracy (B.Acc.), and F1 score. For the three-class FDG-PET-CT-Lesions task, AUC is computed using a one-vs-rest formulation and macro-averaged across classes, while F1 is also macro-averaged. For the binary HECKTOR 2022 recurrence task, the corresponding binary metrics are reported. Segmentation performance is evaluated using the Dice similarity coefficient. All results are reported as mean \(\pm\) standard deviation across the five held-out cross-validation folds.

\subsection{Segmentation Performance}

Although segmentation performance is not the primary focus of this study, we report it to assess the segmentation capability underlying the object pointer representations. Appendix Table~\ref{tab:seg_results} shows that both segmentation-only and task-guided models maintain good segmentation performance across the two datasets. As expected, Dice is slightly higher for the segmentation-only model, while task-guided training introduces a modest reduction as the representation is additionally optimised for downstream classification. Nevertheless, segmentation performance remains largely preserved, with rare segmentation failures observed across both settings.

\begin{figure}[h!]
\centering
\includegraphics[width=1.0\linewidth]{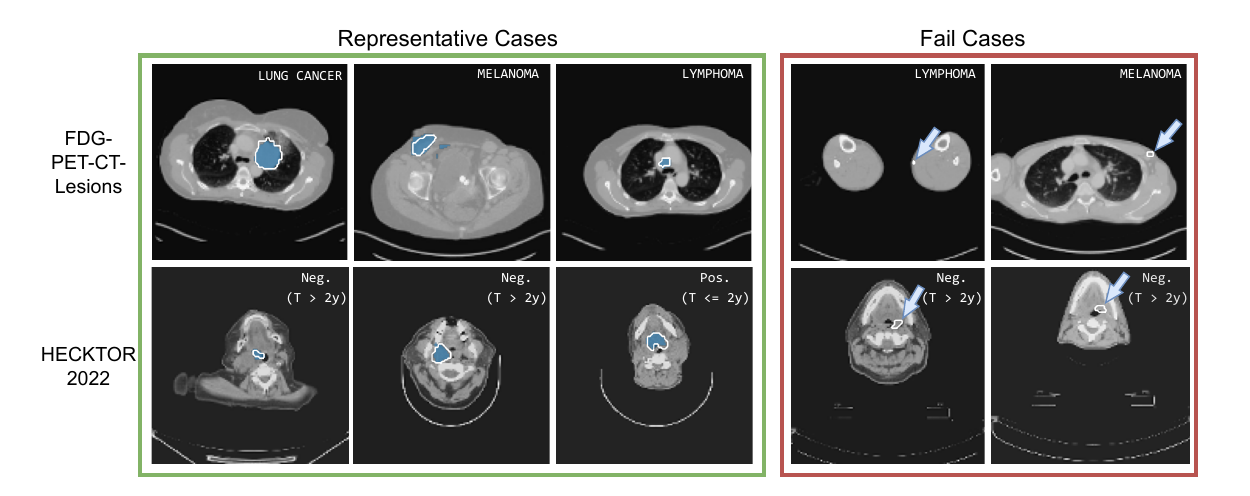}
\caption{Qualitative segmentation results from the segmentation-only object pointer model. Representative successful segmentations (left) and examples of segmentation failures (right) are shown for FDG-PET-CT-Lesions and HECKTOR 2022. White contours indicate the ground-truth tumour annotation and blue regions indicate the predicted segmentation. Arrows highlight the ground-truth tumour location in failure cases. The corresponding downstream class is shown in the top-right corner of each image.}
\label{fig:seg_vis}
\end{figure}

Appendix Fig.~\ref{fig:seg_vis} provides qualitative examples from the segmentation-only model, including representative successful segmentations and failure cases. The failures predominantly occur for very small lesions.

\begin{table}[h!]
\caption{Segmentation performance of object pointer models. Dice is reported as mean \(\pm\) standard deviation across five-fold cross-validation, alongside the total number of segmentation failures.}
\label{tab:seg_results}
\begin{center}
\small
\begin{tabular}{lcc|cc}
\toprule
\multirow{2}{*}{\textbf{Method}}
& \multicolumn{2}{c|}{\textbf{FDG-PET-CT-Lesions}}
& \multicolumn{2}{c}{\textbf{HECKTOR 2022}} \\
\cmidrule(lr){2-3} \cmidrule(l){4-5}
& \textbf{Dice}$\uparrow$
& \textbf{Seg. failures (n)}$\downarrow$
& \textbf{Dice}$\uparrow$
& \textbf{Seg. failures (n)}$\downarrow$\\
\midrule

ObjPtr (Seg.)
& 0.737 $\pm$ 0.013
& 9/489
& 0.693 $\pm$ 0.022
& 3/427 \\

ObjPtr (Task)
& 0.715 $\pm$ 0.010
& 3/489
& 0.685 $\pm$ 0.017
& 0/427 \\
\bottomrule
\end{tabular}
\end{center}
\end{table}

\subsection{PET Normalisation Ablation}
\label{app:pet_normalisation}

We compare three PET intensity normalisation strategies for the segmentation-only object pointer setting on FDG-PET-CT-Lesions. Global normalisation uses the mean and standard deviation estimated from the training dataset and preserves relative intensity variation between subjects and slices. Volume-wise normalisation independently standardises each 3D PET volume using its own intensity statistics, removing inter-patient differences in intensity scale. Slice-wise normalisation independently standardises each axial PET slice, additionally removing slice-to-slice differences within each volume. Table~\ref{tab:norm_results} reports the corresponding segmentation and downstream object pointer performance.

\begin{table}[h!]
\caption{Effect of PET intensity normalisation on segmentation-only object pointer representation quality on FDG-PET-CT-Lesions. Results are reported as mean $\pm$ standard deviation across five-fold cross-validation. The highest mean for each metric is highlighted in \textbf{bold}.}
\label{tab:norm_results}
\begin{center}
\footnotesize
\begin{tabular}{lccc|c}
\toprule
\multirow{2}{*}{\textbf{PET Normalisation}}
& \multicolumn{3}{c|}{\textbf{Classification}}
& \multicolumn{1}{c}{\textbf{Segmentation}} \\
\cmidrule(lr){2-4} \cmidrule(l){5-5}
& \textbf{AUC}$\uparrow$
& \textbf{B.Acc.}$\uparrow$
& \textbf{F1}$\uparrow$
& \textbf{Dice}$\uparrow$ \\

\midrule
Global normalisation
& \textbf{0.781 $\pm$ 0.057}
& \textbf{0.611 $\pm$ 0.075}
& \textbf{0.604 $\pm$ 0.072}
& \textbf{0.737 $\pm$ 0.013} \\

3D volume-wise normalisation
& 0.762 $\pm$ 0.044
& 0.567 $\pm$ 0.059
& 0.560 $\pm$ 0.056
& 0.728 $\pm$ 0.022 \\

2D slice-wise normalisation
& 0.723 $\pm$ 0.059
& 0.525 $\pm$ 0.067
& 0.519 $\pm$ 0.060
& 0.703 $\pm$ 0.014 \\

\bottomrule
\end{tabular}
\end{center}
\end{table}

\subsection{Object Pointer-Space Visualisation}
\label{app:obj_ptr_space}

\begin{figure}[!h]
\centering
\includegraphics[width=1.0\linewidth]{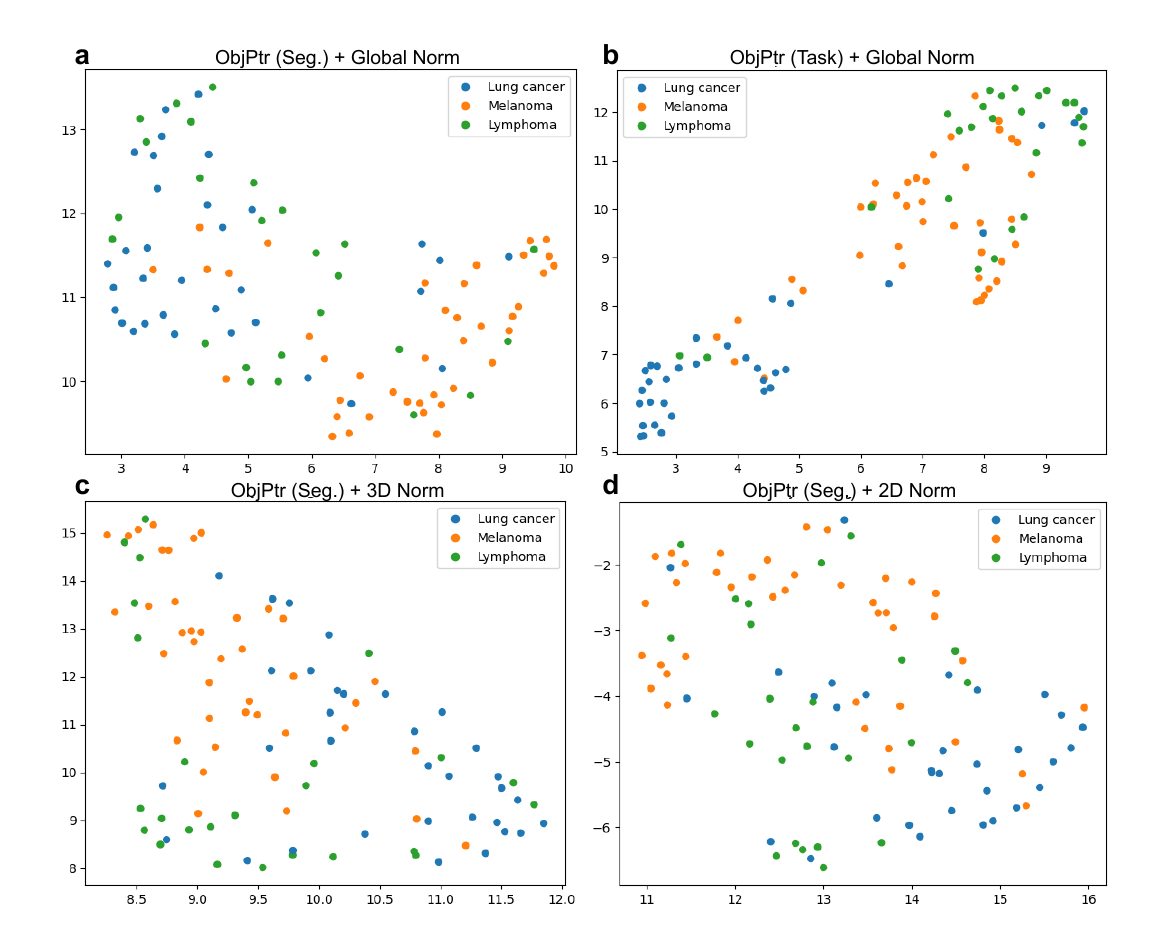}
\caption{Organisation of the SAM2 object pointer representation space on FDG-PET-CT-Lesions. UMAP projections of tumour-level object pointers for (a) the segmentation-only model with global PET norm, (b) task-guided object pointer training, (c) segmentation-only 3D volume-wise, and (d) segmentation-only 2D slice-wise PET normalisation. Points represent individual tumours and are coloured by cancer type. The segmentation-only representation already exhibits class-dependent organisation, while changes in PET normalisation alter this structure and task-guided training further separates the downstream classes.}
\label{fig:clustering}
\end{figure}

To qualitatively examine the organisation of the learned object pointer space on FDG-PET-CT-Lesions, we project the 256-dimensional tumour-level representations into two dimensions using UMAP~\citep{mcinnes2018umap}. Each point represents a subject and is coloured according to cancer type. Fig.~\ref{fig:clustering} shows a representative fold, comparing segmentation-only and task-guided object pointers, as well as segmentation-only representations obtained under different PET normalisation strategies. UMAP is fitted independently to each representation using 15 nearest neighbours and cosine distance.

\subsection{Radiomics Association Analysis}

\subsubsection{Canonical correlation analysis}
Canonical correlation analysis (CCA) is used to assess multivariate associations between the tumour-level object pointer representation and radiomics features. Radiomics features are grouped according to their respective feature families, including shape, first-order intensity, and texture-based features. Within each cross-validation fold, CCA is fitted using the training subjects and the learned projections are applied to the corresponding held-out test subjects. We report the first canonical correlation, corresponding to the strongest linear association between the two projected feature spaces.

\subsubsection{Pearson correlation analysis}
\begin{figure}[h!]
\centering
\includegraphics[width=1.0\linewidth]{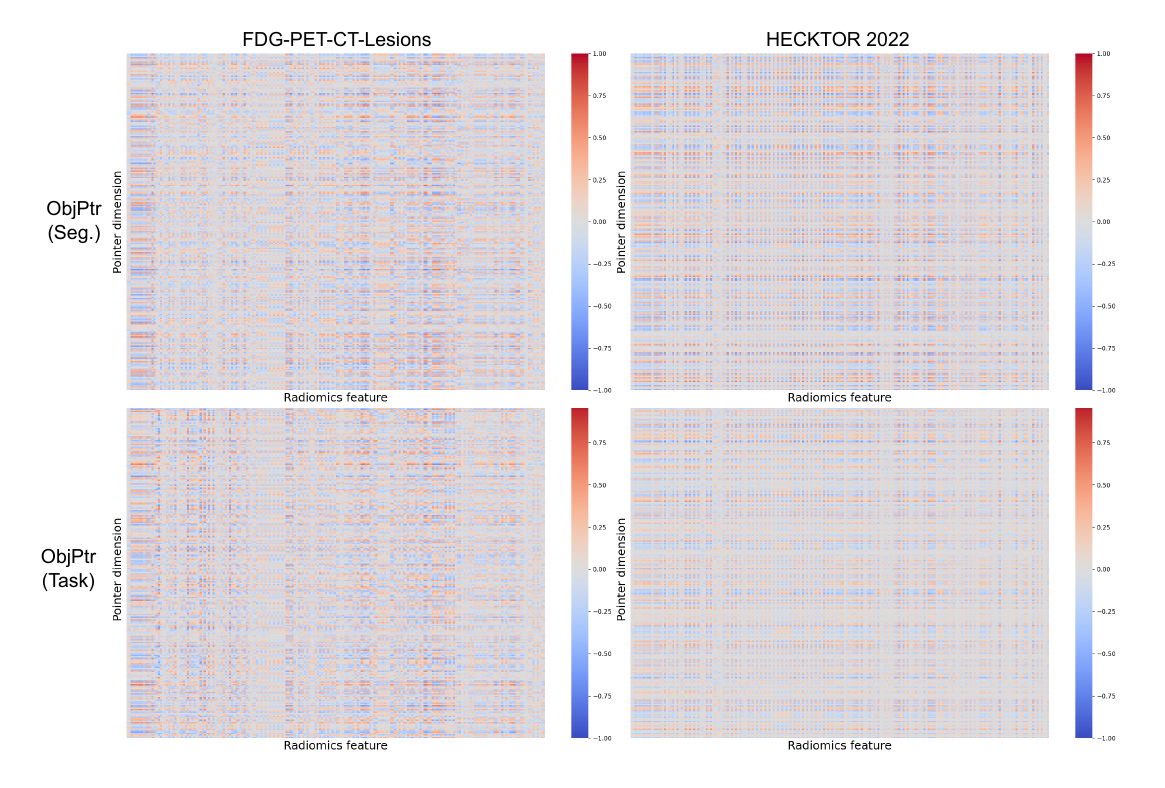}
\caption{Pearson correlation coefficients (PCC) between individual object pointer dimensions and radiomics features. PCC is shown for segmentation-only and task-guided object pointers across FDG-PET-CT-Lesions and HECKTOR 2022.}
\label{fig:pcc_heat}
\end{figure}

To complement the multivariate CCA analysis, we additionally examine associations between individual object pointer dimensions and individual radiomics features using Pearson correlation coefficients (PCC). Correlations are computed between each of the 256 object pointer dimensions and each extracted radiomics feature. Fig.~\ref{fig:pcc_heat} shows the resulting correlation matrices for segmentation-only and task-guided object pointer representations across both datasets.

\subsection{Auxiliary Loss Ablation}
\label{app:aux_ablation}

\begin{figure}[h!]
\centering
\includegraphics[width=1.0\linewidth]{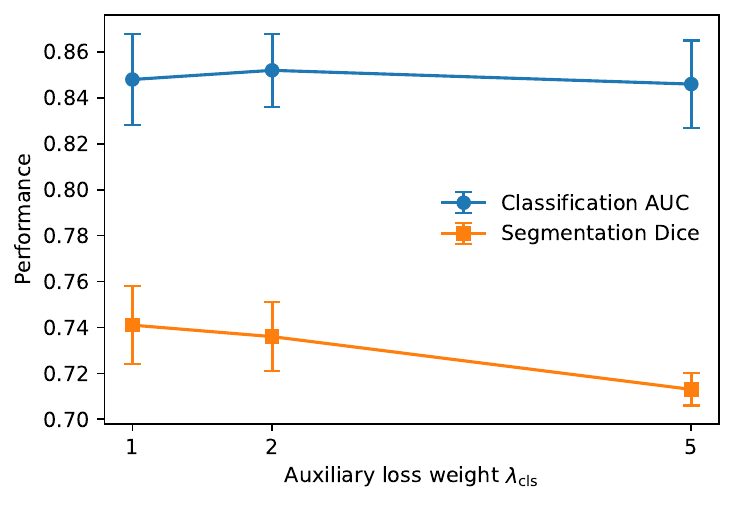}
\caption{Effect of auxiliary classification-loss weight on FDG-PET-CT-Lesions. Downstream classification AUC and segmentation Dice across different values of \(\lambda_{\mathrm{cls}}\). Points show the mean across five-fold cross-validation, with error bars indicating standard deviation.}
\label{fig:reg_ablation}
\end{figure}

We evaluate $\lambda_{\mathrm{cls}} \in \{1, 2, 5\}$ to assess the effect of increasing the contribution of the auxiliary classification objective on FDG-PET-CT-Lesions.  Fig.~\ref{fig:reg_ablation} reports downstream AUC and segmentation Dice for each setting on the validation set. 

\end{document}